\documentclass{article}
\usepackage{ijcai22}

\usepackage{times}
\usepackage{soul}
\usepackage{url}
\usepackage[hidelinks]{hyperref}
\usepackage[utf8]{inputenc}
\usepackage[small]{caption}
\usepackage{graphicx}
\usepackage{amsmath}
\usepackage{amsthm}
\usepackage{booktabs}
\usepackage[ruled,vlined]{algorithm2e}
\usepackage{amsfonts,amssymb}
\usepackage{latexsym}
\usepackage[T1]{fontenc}
\usepackage[utf8]{inputenc}
\usepackage{microtype}
\usepackage{multirow}
\usepackage{colortbl}
\usepackage{subfigure}
\usepackage{diagbox}

\usepackage[ruled,vlined]{algorithm2e}
\usepackage{amsfonts,amssymb}

\title{Searching for Optimal Subword Tokenization in Cross-domain NER}

\author{
Ruotian Ma\textsuperscript{\rm 1}\thanks{\ \ Equal contribution.}
\and
Yiding Tan\textsuperscript{\rm 1}\footnotemark[1]\and
Xin Zhou\textsuperscript{\rm 1}\and
Xuanting Chen\textsuperscript{\rm 1}\and
Di Liang\textsuperscript{\rm 3}\and \\
Sirui Wang\textsuperscript{\rm 3}\and
Wei Wu\textsuperscript{\rm 3} \and
Tao Gui\textsuperscript{\rm 2}\thanks{\ \  Corresponding author.}\and 
Qi Zhang\textsuperscript{\rm 1}
\affiliations
$^1$School of Computer Science, Fudan University, Shanghai, China\\
$^2$Institute of Modern Languages and Linguistics, Fudan University, Shanghai, China\\
$^3$Meituan Inc., Beijing, China
\emails
\{rtma19,yidingtan20,tgui\}@fudan.edu.cn
}

\begin{document}

\maketitle

\begin{abstract}

Input distribution shift is one of the vital problems in unsupervised domain adaptation (UDA). The most popular UDA approaches focus on domain-invariant representation learning, trying to align the features from different domains into similar feature distributions. However, these approaches ignore the direct alignment of input word distributions between domains, which is a vital factor in word-level classification tasks such as cross-domain NER. 
In this work, we shed new light on cross-domain NER by introducing a subword-level solution, X-Piece, for 
input word-level distribution shift in NER. Specifically, we re-tokenize the input words of the source domain to approach the target subword distribution, which is formulated and solved as an optimal transport problem. As this approach focuses on the input level, it can also be combined with previous DIRL methods for further improvement. Experimental results show the effectiveness of the proposed method based on BERT-tagger on four benchmark NER datasets. Also, the proposed method is proved to benefit DIRL methods such as DANN.
\end{abstract}

\section{Introduction}
Unsupervised domain adaptation (UDA) is a widely concerned problem in Machine Learning \cite{pan2010domain,ganin2016domain}, which is also a significant point in Named Entity Recognition (NER) \cite{chen2019transfer}. The main concern in UDA is to perform a given task on a target domain with only unlabeled data provided, while having rich labeled data on a source domain. A key problem is that the data distribution varies across different domains, resulting in poor performance when directly adopting a source domain model to the target domain. 

The most popular line of UDA researches focus on domain invariant representation learning (DIRL) \cite{pan2010domain,ganin2016domain}.
Given that the input distribution shift exists between domains, these methods attempt to learn similar feature distributions for both source and target domains,
thus a classifier trained on the source domain can easily generalize to the target domain. Recent studies have verified the effectiveness of DIRL methods on several text classification such as sentiment analysis \cite{pengetal2018cross}, duplicate question detection \cite{shahetal2018adversarial} and relation extraction \cite{fuetal2017domain}. In most cases, deep features in neural networks would transition from general to specific along the network, and the transferability of features and classifiers decreases as the cross-domain discrepancy increases \cite{NIPS2014_375c7134,pmlr-v70-long17a}. 

\begin{figure}
    \centering
    \includegraphics[width=0.86\linewidth]{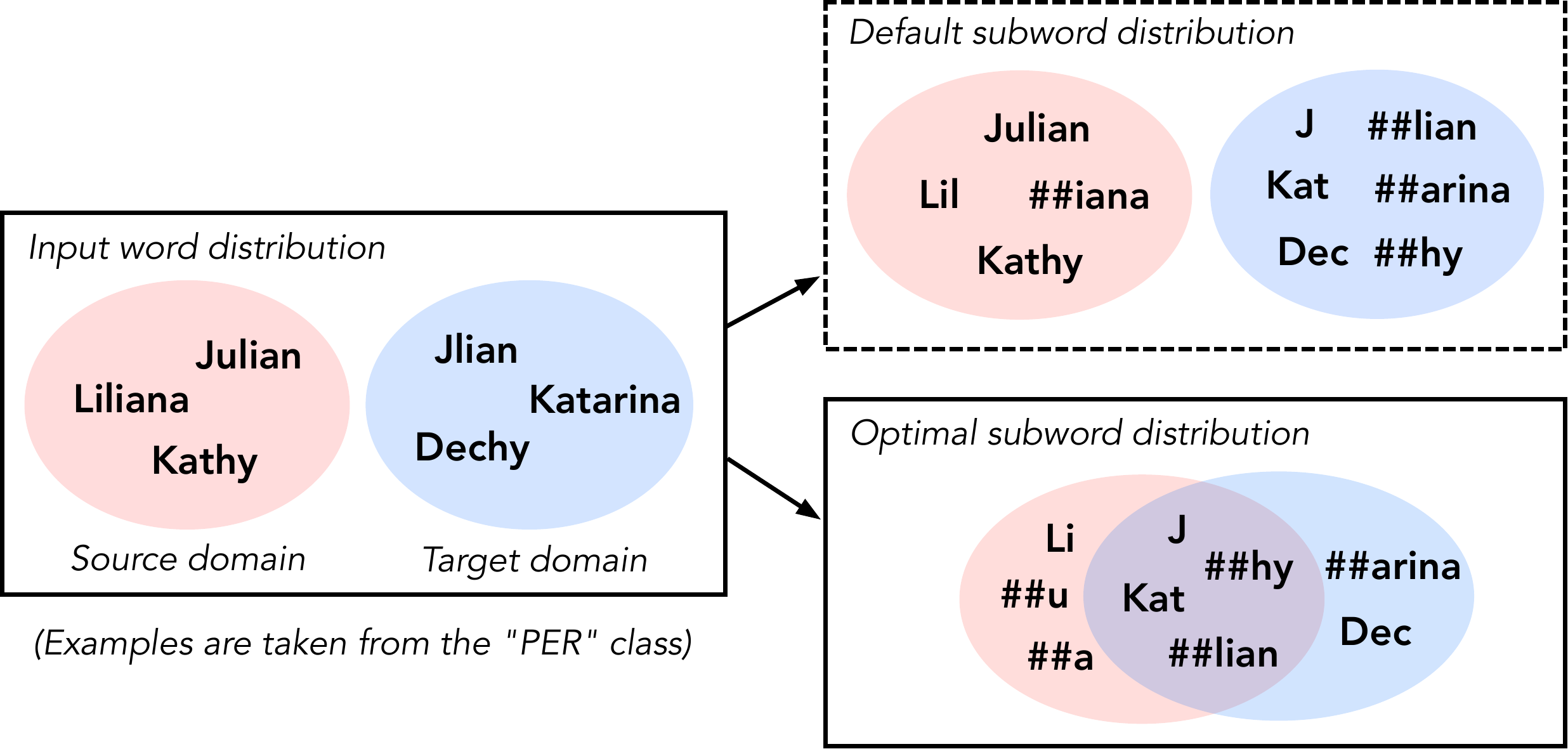}
    \caption{An intuitive overview of the proposed method. As seen, there is a large discrepancy between the source and target word distributions, as well as the default subword distributions. However, after altering the subword segmentation of the source domain, the subword distributions get much closer.}
    \label{fig:example}
    \vspace{-0.3cm}
\end{figure}

However, previous methods mainly focus on transferring high-dimensional features, while having ignored alleviating the domain discrepancy from the data perspective. Especially, such feature-based methods may have more difficulties for word-level classification tasks such as NER that largely depends on the input word-level distribution.
\cite{fu2020rethinking} has highlighted that the domain discrepancy of input word-label distribution leads to poor cross-domain performance even with the powerful transferability of the pre-trained LM representations. For instance, the word ``Madrid" might be annotated with different labels across domains (``Location" in a news domain and ``Organization" when it appears as ``Real Madrid Club" in a sport domain), thus the source domain model might predict a wrong label in the target domain. \cite{jiazhang2020multi} also pointed out that the main concerns of cross-domain NER fall into the inconsistence of cross-domain word-label distributions and the existence of new entity words.
Hence, how to alleviate the input discrepancy when the word distribution of datasets is determined is an important issue.

In this work, we shed new light on cross-domain NER by introducing a subword-level solution, X-Piece, to directly alleviate the input word-level distribution shift. As shown in Figure \ref{fig:example}, we re-tokenize the source domain input words in order to find an optimal subword distribution that approaches the target subword distribution. To optimize the discrete subword distribution, we formulate it as an optimal transport problem, which is solved to minimize the KL-divergence of the source and target conditional distribution ($P(label|subword)$). To estimate the target conditional distribution, we use the entity lexicon to obtain distant annotations on the unlabeled target data. Such annotation is noisy since it might mistake a large number of entities that do not exist in the lexicon. However, our approach is relatively robust to the noise as it depends only on the distribution, instead of directly training on noisy labels.
Additionally, as an input-level solution, X-Piece is complementary to DIRL methods that deal with the feature-level distributions.
Experimental results show that the proposed method X-Piece obtains consistent gains based on BERT-tagger over different datasets. Also, the proposed method is shown to benefit the DIRL methods such as DANN.\footnote{Our code is available at \textit{https://github.com/rtmaww/X-Piece}.}

To summarize the contribution of this work:
\begin{itemize}
\setlength{\itemindent}{0em}
\setlength{\itemsep}{0em}
\setlength{\topsep}{-0.5em}
    \item We present a first attempt to solve the domain adaptation problem via a subword-level distribution alignment.
    \item We propose X-Piece, a novel method based on optimal transport to re-tokenize the source domain inputs for alleviating distribution shift.
    \item Experimental results show the effectiveness of X-Piece based on BERT-tagger on four NER datasets.
\end{itemize}

\section{Related Works}

\paragraph{Unsupervised Domain Adaptation}
Unsupervised domain adaptation (UDA) is a widely concerned problem in machine learning, whose target is to train a model that performs well on a target domain test set with only unlabeled data in the target domain and sufficient labeled data in the source domain. Typical UDA researches include: (1) the feature-based methods \cite{pan2010domain,ganin2016domain} which attempts to learn a feature encoder that maps the source and target inputs into similar feature distributions, thus a classifier trained on the source domain can easily generalize to the target domain. However, as these methods manipulate on the high-dimensional feature distribution instead of directly solving the input data distribution shift, they are less suitable for word-level classification tasks that largely relates to word-level distribution. (2) the instance-based methods \cite{jiang-zhai-2007-instance,cortes2008sample} that leverage the weights of the input instances in order to alleviate the input distribution shift. These methods are similar to our method. However, we directly change the input distribution in a subword-level instead of softly weighting, which is more effective.

\paragraph{Domain Adaptation for NER}
Several researches have studied cross-domain NER \cite{chen2019transfer,Liu_Xu_Yu_Dai_Ji_Cahyawijaya_Madotto_Fung_2021}. These methods mainly focuses on feature-based \cite{jiazhang2020multi} or data-based \cite{zhang-etal-2021-crowdsourcing} domain adaptation, while our approach is completely different from these methods as we are the first to deal with the subword-level distribution for domain adaptation.

\begin{figure*}
    \centering
    \includegraphics[width=0.88\linewidth]{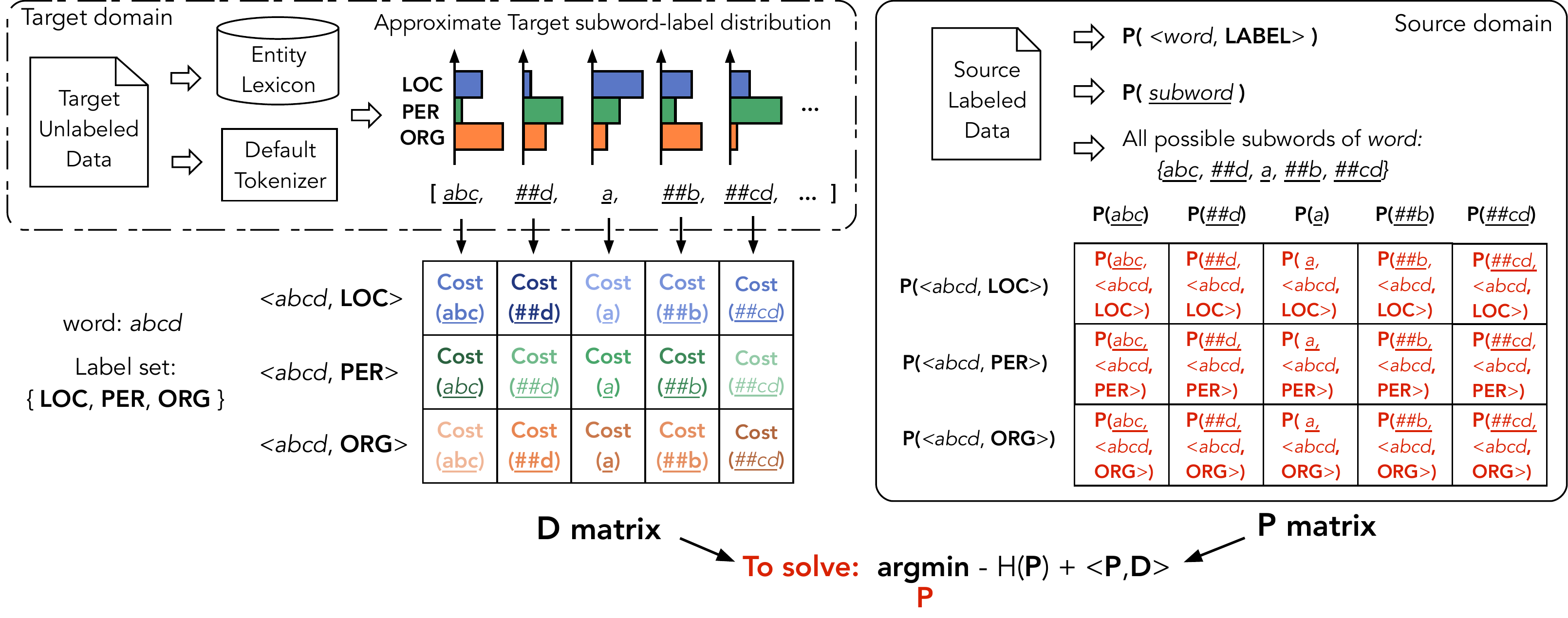}
    \caption{An overview of the optimal transport process in the proposed method, X-Piece. The OT aims to solve the transporting mass from the subword distribution to the word-label distribution with a minimal cost. Here, the cost matrix (or distance matrix) is obtained from the target subword-label distribution, where the cost increases with lower $P_T(y|{t})$. After solving the OT problem, we can obtain an optimal subword-word-label distribution $P_S(\mathbf{t},\mathbf{w},\mathbf{y})$ in the source domain, which is to guide the re-tokenization of the source domain inputs.}
    \label{fig:OT_main}
    \vspace{-0.2cm}
\end{figure*}

\paragraph{Subword Segmentation in NLP}
Subword segmentation has become a standard process in NLP, which segments input words into subword units to alleviate the problem of infrequent words with a acceptable vocabulary size \cite{devlin-etal-2019-bert,liu2019roberta}.
However, typical segmentation approach, such as the WordPiece Tokenizer used in BERT \cite{devlin-etal-2019-bert}, splits words into unique sequences with a greedy longest-match-first algorithm, only considering the frequency and length of subwords for word segmentation. Such approach fails to perform various and task-specific tokenization on downstream tasks.
To improve the problem, \cite{kudo-2018-subword} and \cite{provilkov-etal-2020-bpe} propose regularization methods to allow multiple segmentations for a word, while they do not fix the domain shift problem. 
\cite{schick-schutze-2020-bertram,liu-etal-2021-bridging} propose methods to generate embeddings for the rare words and the open-vocabulary words under pretrain-finetune domain shift. Similar to our approach, \cite{sato-etal-2020-vocabulary} focuses on domain adaptation and adapts the embedding layers of a given model to the target domain.
Generally, these DA methods are label-agnostic embedding-level methods trying to alleviate the problem of different marginal distribution $P_S(\mathbf{x})\neq P_T(\mathbf{x})$ between domains. While our method aims at aligning the conditional or joint distribution ($P_S(\mathbf{y}|\mathbf{x})=P_T(\mathbf{y}|\mathbf{x})$ or $P_S(\mathbf{x},\mathbf{y})=P_T(\mathbf{x},\mathbf{y})$) under domain shift from a subword segmentation perspective.

\section{Approach}
In this section, we introduce a subword-level domain adaptation approach, X-Piece, for cross-domain NER. 

\subsection{Problem Statement}
In this work, we study cross-domain NER following the unsupervised domain adaptation setting. Specifically, we assume a source domain $\mathcal{D}_S$ and a target domain $\mathcal{D}_T$ with different distribution over $X \times Y$. 
We are provided with a labeled dataset drawn from the source domain $S=\{{X}^i,{Y}^i\}^n_{i=1}\sim \mathcal{D}_S$ and an unlabeled dataset drawn from the target domain $T=\{{X}^i\}^{n'}_{i=1} \sim \mathcal{D}_T$, where $X^i=\{x_1^i,\dots, x_m^i\}$ and $Y^i=\{y_1^i,\dots, y_m^i\}$ are the input word sequence and corresponding labels of the $i^{th}$ input sample. DA assumes that the two domains have different marginal distribution $P_S(\textbf{x}) \neq P_T(\textbf{x})$ and different conditional distribution $P_S(\textbf{y}|\textbf{x}) \neq P_T(\textbf{y}|\textbf{x})$, thus have different joint distribution $P_S(\textbf{x},\textbf{y}) \neq P_T(\textbf{x},\textbf{y})$. Then, the aim of cross-domain NER is to train a classifier $f: {X} \to {Y}$ using $
S$ and $T$ that performs well on the target domain test set $T_{test}$.

\subsection{Domain Adaptation via Optimal Word Tokenization}

To alleviate the domain shift in the input-level, we propose to align the subword distribution in the source domain towards the target distribution via re-tokenizing the input words. We formulate and solve this discrete optimization problem as an optimal transport problem. An overview of the proposed method is shown in Fig.\ref{fig:OT_main}.

\subsubsection{KL-divergence as Optimization Objective}\label{KLdivergence}
Formally, we assume an identical subword space $\mathbb{T}$ and a default tokenizer $\mathcal{T}_{ori}$ in both domains. Given any input word $x$, the default tokenizer tokenize $x$ into a subword sequence $\mathcal{T}_{ori}(x)=\{t_1,\dots,t_k\}, t_i\in \mathbb{T}$. Typical tokenizers used in pre-trained LM, such as the Wordpiece tokenizer in BERT-tagger, is usually an injection function, meaning that an input word $x$ is always tokenized into one identical subword sequence regardless of the context and the label. Therefore, we also have $P_{S,ori}(\textbf{t},\textbf{y}) \neq P_{T,ori}(\textbf{t},\textbf{y})$ and $P_{S,ori}(\textbf{y}|\textbf{t}) \neq P_{T,ori}(\textbf{y}|\textbf{t})$.

Consequently, to solve the distribution shift problem across domains, we are to search for an optimal tokenizer $\mathcal{T}^*$ in the source domain such that $P_{S,*}(\textbf{t},\textbf{y}) \approx P_{T,ori}(\textbf{t},\textbf{y})$ (joint) or $P_{S,*}(\textbf{y}|\textbf{t}) \approx P_{T,ori}(\textbf{y}|\textbf{t})$ (conditional) . Here, we calculate the KL-divergence score as the measurement of distribution distance. The problem can then be formulated as:

\begin{equation}
\setlength\abovedisplayskip{-1pt}
\setlength\belowdisplayskip{-1pt}
\begin{aligned}
\mathcal{T}^* = \arg\min_{\mathcal{T}}D_{KL}(P_S\|P_T)
\end{aligned}
\end{equation}
where $P$ denotes the joint or conditional distribution.

\subsubsection{Formulating Optimal Transport (OT) Problem}\label{formulate}

When searching for the optimal tokenizer $\mathcal{T}^*$, we are actually searching for an optimal joint distribution $P_S^*(\textbf{t},\textbf{w},\textbf{y})$ of subwords, words and labels, and this joint distribution is to guide the tokenization of each word under a certain label. As a result, this tokenization process can be regarded as transporting the subword distribution $P_S(\textbf{t})$ into the word-label distribution $P_S(\textbf{w},\textbf{y})$. Therefore, we can formulate it as a \textbf{discrete optimal transport problem} with a cost considering the source-target discrepancy, which can then be solved efficiently via the Sinkhorn algorithm.

Formally, we reformulate the KL-divergence objective as the following objective function which has the same form as the objective function in optimal transport (The detailed derivation can be found in Appendix):
\begin{equation}\label{OT_final}
\begin{aligned}
\min_{\mathbf{P}\in \mathcal{R}^{(w,y)\times t}}<\mathbf{P},\mathbf{D}>-\gamma \mathcal{H}(\mathbf{P})
\end{aligned}
\end{equation}
Here, $\mathbf{P}$ is exactly the optimal joint distribution $P_S(\textbf{t},\textbf{w},\textbf{y})$ we search for, which is then to guide the optimal tokenization $\mathcal{T}^*$. $\mathbf{D}$ is the transporting cost matrix related to the distribution distance between domains, which has different form when optimizing based on the joint and conditional distribution:
\begin{equation}
\label{eq6}
D(t,w,y)=\left\{
\begin{aligned}
&-log(\frac{P_T(t,y)}{P_T(t)}), &Conditional \\
&-log(P_T(t,y)), &Joint
\end{aligned}
\right.
\end{equation}

\subsection{Setup of Optimal Transport}
In the above formulated OT problem, we are to find the best transporting mass from the subword distribution to the word-label distribution. Here we further ensure the validation of the transport solutions by adding some constraints.

First, we set the initial values of each $(w,y)$ term and each $t$ term based on the word-label distribution in the source domain:
\begin{equation}
\setlength\abovedisplayskip{3pt}
\setlength\belowdisplayskip{0pt}
\begin{aligned}
&\mathcal{A}(w,y)=\phi(w,y)\cdot |sub(w)| \\
&\mathcal{A}(t)=\sum_w \phi(w)\cdot\mathcal{I}[t\in sub(w)]
\end{aligned}
\end{equation}
where $\phi(w,y)$ and $\phi(w)$ denotes the frequency of the word-label pair $(w,y)$ and the frequency of the word $w$ itself in the source data. $sub(\cdot)$ is the set of all the possible subwords segmented from the word $w$. $\mathcal{I}[t\in sub(w)]$ is a indicator function that stays true if the subword $t$ exists in $sub(w)$.

Next, considering that the word-label distribution and the possible subwords that can be tokenized from a certain word is fixed, we add the constraints to keep the distributions of each row and column:
\begin{equation}
\setlength\abovedisplayskip{3pt}
\setlength\belowdisplayskip{0pt}
\begin{aligned}
\sum_{(w,y)}{P}(t,w,y)&=\mathcal{A}(t), 
\sum_t{P}(t,w,y)&=\mathcal{A}(w,y)
\end{aligned}
\end{equation}

Then, we fix the cost matrix $\textbf{D}$ considering that a subword should not transport to a word that does not cover it:
\begin{equation}
\setlength\abovedisplayskip{3pt}
\setlength\belowdisplayskip{0pt}
\begin{aligned}
\hat{D}(t,w,y)=
\left\{
\begin{array}{lr}
      {D}(t,w,y), & if \quad t \in sub(w) \\
      +\infty, & if \quad  t \notin sub(w) 
\end{array}
\right.
\end{aligned}
\end{equation}

As a result, we can solve the above optimal transport problem through the Sinkhorn algorithm. The details can be found in Appendix.

\begin{table}\label{ontonotes}
\centering
\small

\begin{tabular}{lcccc}
\toprule
\textbf{Datasets} & \textbf{Domain} & \textbf{\# Class} & \textbf{\# Train} & \textbf{\# Test} \\
\midrule
CoNLL'03 & News & 3 & 14.0k & 3.5k \\
OntoNotes  & General & 18 & 60.0k & 8.3k\\
Twitter & Social Media & 3 & 2.3k & 3.8k\\
Webpages & Website & 3 & 385 & 135\\
\bottomrule

\end{tabular}
\caption{Dataset details. Here we only show the actual class number we use in cross-domain settings.}
\label{tab:dataset}
\vspace{-0.9em}
\end{table}

\subsubsection{Target Distribution Approximation with Lexicon}\label{lexicon}
As we have finished formulating the optimal transport problem, one remaining problem is that the target domain label distribution is not accessible in the unsupervised domain adaptation setting. However, in NER task, it is easy to access large number of entity lexicons, which can be used as a distant annotation on the unlabeled data for approximating the target distribution. Specifically, we use 
the KB-based lexicon annotation algorithm proposed by \cite{liang2020bond} to annotate the unlabeled data. Then, we use the default tokenizer $\mathcal{T}_{ori}$ to tokenize the target data, in order to obtain the subword and subword-label distributions we need to calculate the cost term $log(\frac{P_T(t,y)}{P_T(t)})$ or $log(P_T(t,y))$. Note that the lexicon annotation is inevitably noisy since the lexicon cannot cover all of the entities in the target data. Such unlabeled entity problem will 
cause a performance degradation if directly used for training \cite{li2021empirical}. However, our approach only estimates the target distribution and do not directly train on the noisy data, thus is relatively robust to the noisy annotation, which is also proved by the experimental results.

\renewcommand\arraystretch{1.0}
\begin{table*}[ht]
\centering
\small

\begin{tabular}{c|l|p{1cm}<{\centering}|p{1cm}<{\centering}|p{1cm}<{\centering}|p{1cm}<{\centering}|p{1cm}<{\centering}|p{1cm}<{\centering}}
\hline
\hline
\multicolumn{1}{c|}{\multirow{1}{*}{\textbf{Source}}} &
\multicolumn{1}{c|}{\multirow{2}{*}{\textbf{Methods}}} & 
\multicolumn{6}{c}{\multirow{1}{*}{\textbf{Target domains}}}   \\
\cline{3-8}
\textbf{domains} & &
\multicolumn{1}{c|}{\multirow{1}{*}{\textbf{BC}}} & 
\multicolumn{1}{c|}{\multirow{1}{*}{\textbf{BN}}} & 
\multicolumn{1}{c|}{\multirow{1}{*}{\textbf{MZ}}} & 
\multicolumn{1}{c|}{\multirow{1}{*}{\textbf{NW}}} &
\multicolumn{1}{c|}{\multirow{1}{*}{\textbf{TC}}} &
\multicolumn{1}{c}{\multirow{1}{*}{\textbf{WB}}} \\
\hline
\multirow{2}{*}{\textbf{BC}}
&

BERT-tagger &  - & 82.90  & \textbf{80.57}  & 76.28 & 69.42 & \textbf{73.61} \\ 
&\textbf{X-Piece} & - & \textbf{83.33}  & 80.53 & \textbf{76.59} & \textbf{69.96} & 72.48 \\

\hline

\multirow{2}{*}{\textbf{BN}}&

BERT-tagger &  79.26  &  - & 79.07  & 81.78 & 71.20 & 74.39\\ 
&\textbf{X-Piece} & \textbf{79.35} &  - & \textbf{79.88} & \textbf{82.97} & \textbf{71.46} & \textbf{74.49} \\

\hline

\multirow{2}{*}{\textbf{MZ}}&

BERT-tagger &  72.76  & \textbf{80.04}  & -  & 79.03 & \textbf{70.85} & 75.04 \\ 
&\textbf{X-Piece} & \textbf{74.52} & {79.79}  & - & \textbf{80.60} & 68.32& \textbf{77.33}\\

\hline

\multirow{2}{*}{\textbf{NW}}&

BERT-tagger &  78.65  & 86.62  & 83.85  & - & 70.71 & 75.07 \\ 
&\textbf{X-Piece} & \textbf{79.09} & \textbf{87.37}  & \textbf{84.70} & - &\textbf{70.91} & \textbf{76.24} \\

\hline

\multirow{2}{*}{\textbf{TC}}&

BERT-tagger &  54.68  & 63.85  & 50.09  & 52.29 & - & \textbf{61.94}\\ 
&\textbf{X-Piece} & \textbf{55.80} & \textbf{64.92}  & \textbf{54.67} & \textbf{53.55} & - & {61.57}\\

\hline

\multirow{2}{*}{\textbf{WB}}&

BERT-tagger &  69.60  & 77.06  & \textbf{78.51}  & 74.63 & 66.33 & - \\ 
&\textbf{X-Piece} & \textbf{70.85} & \textbf{78.79}  & 78.46 & \textbf{75.69} & \textbf{67.28} & -\\

\hline
\hline
\end{tabular}
\caption{Cross-domain results (F1 score) of X-Piece and BERT-tagger on Ontonotes cross-domain setting. Here, BC, BN, MZ, NW, TC and WB are six different sub-domains in the Ontonotes 5.0 dataset.}
\label{tab:ontonotes}
\end{table*}

\begin{table*}[ht]
\centering
\small
\begin{tabular}{c|l|p{1.5cm}<{\centering}|p{1.5cm}<{\centering}|p{1.5cm}<{\centering}}
\hline
\hline
\multicolumn{1}{c|}{\textbf{Datasets}} &
\multicolumn{1}{c|}{\textbf{Methods}} & 
\multicolumn{1}{c|}{\textbf{OntoNotes 5.0}} & 
\multicolumn{1}{c|}{\textbf{Webpages}} & 
\multicolumn{1}{c}{\textbf{Twitter}}
  \\ \cline{1-5}
\hline
\multirow{6}{*}{\textbf{CoNLL'03}}

& BERT-tagger \cite{devlin-etal-2019-bert} &  78.33  & 74.61  & 53.26  \\ 
& \textbf{X-Piece} & \textbf{78.91} & \textbf{78.12}  & \textbf{53.43} \\
\cline{2-5}
& Multi-cell LSTM \cite{jiazhang2020multi}&  63.07  & 72.39  & 44.50 \\ 
& BERT+CMD \cite{zellinger2017central}&  78.90  & 74.01  & 53.10   \\ 
& BERT+EADA \cite{zou-etal-2021-unsupervised}& 77.84 & 78.12 & 53.29 \\
& BERT+DANN \cite{ganin2016domain}&  {79.04}  & 78.75  & 54.07   \\
& \textbf{X-Piece + DANN} & \textbf{79.54} & \textbf{79.99}  & \textbf{54.17}  \\


\hline
\hline
\end{tabular}
\caption{Cross-domain results (F1 score) of X-Piece and competitive baselines from CoNLL'03 to other datasets.}
\label{tab:conll}
\vspace{-1.0em}
\end{table*}

\subsection{Re-tokenizing Input Words with OT Solution}
After obtaining a solution of the distribution $P(t,w,y)$ via OT, we are to re-tokenize the source domain input words based on $P(t,w,y)$. However, given the subword distribution $P(\textbf{t}|w,y)$ of each word-label pair, there is still a gap between the segmented subword sequence and the distribution. Here, we denote a certain segmentation of word $w$ as $s=\{t_1,\dots,t_k\}$. Calculating the distribution of each segmentation $P(\textbf{s}|w,y)$ based on $P(\textbf{t}|w,y)$ can be regarded as solving a system of linear equations, where for each $t_i \in sub(w)$:

\begin{equation}
\setlength\abovedisplayskip{0pt}
\setlength\belowdisplayskip{0pt}
\begin{aligned}
P(s_1)\cdot C_{s_1,t_i}+\dots+P(s_k)\cdot C_{s_k,t_i}=P(t_i|w,y)
\end{aligned}
\end{equation}
where $C_{s_k,t_i}$ denotes the number of subword $t_i$ that contained by the segmentation $s_k$.

As the coefficient matrix of these linear equations are sparse, we simply assume that each segmentation $s_j$ include a singular subword $t_i$ that does not appear in other segmentation, and take the value $P(t_i)$ as $P(s_j)$ by:
\begin{equation}
\setlength\abovedisplayskip{3pt}
\setlength\belowdisplayskip{0pt}
\begin{aligned}
P(s_j)=\min_{t_i}\{P(t_i|w,y), t_i \in s_j\}
\end{aligned}
\end{equation}

Then, given a word $w$ with label $y$ in the source domain, we re-tokenize it by sampling a segmentation based on $P(\textbf{s}|w,y)$ instead of consistently choosing the highest one, in order to match the $P(t,w,y)$ distribution we have solved:
\begin{equation}
\setlength\abovedisplayskip{3pt}
\setlength\belowdisplayskip{0pt}
\begin{aligned}
\mathcal{T}^*(w|y)=Sample(s) \sim P(\textbf{s}|w,y)
\end{aligned}
\end{equation}

\section{Experiments}

In this work, we conduct experiments on cross-domain NER following the commonly adopted unsupervised domain adaptation (UDA) setting. Specifically, we assume an identical label space of both source and target domains. We use the source domain labeled data and the target domain unlabeled data for model training, and then compare the performance on the target domain test set.

We conduct experiments on four commonly used NER datasets from different domains: CoNLL'03 from the newswire domain, Twitter from the social media domain, Webpages from the web domain, and OntoNotes 5.0 which contains text documents from six domains, including broadcast conversation (BC), broadcast news (BN), magazine (MZ), newswire (NW), web (WB) and telephone conversation (TC). The training set size of these sub-domains are 10.4k, 9.7k, 6.9k, 15.2k, 11.1k, 6.4k, respectively. More details of the datasets are shown in Table \ref{tab:dataset}.

To fully illustrate the proposed method, we adopt two experimental settings: (1) Cross-domain between the OntoNotes sub-domains. (2) Cross-domain from CoNLL'03 to other datasets. To ensure the same label space, when conducting experiments on Setting (2), we consider only PER, LOC, ORG entity classes (For OntoNotes, there are several fine-grained entity types such as NORP that can be included by LOC, thus we merge these entity types as LOC). When conducting experiments on Setting (1), we consider all the 18 entity classes.

\subsection{Baselines}
We include several competitive baselines to verify the effectiveness of the proposed method:

\textbf{BERT-tagger} \cite{devlin-etal-2019-bert} The BERT-based baseline for NER which replaces the LM head of pre-trained BERT model with a token classification head. We use the implementation of Huggingface\footnote{https://github.com/huggingface/transformers} based on the bert-base-cased pre-trained model (also for all the other baselines).

\textbf{DANN} \cite{ganin2016domain} and \textbf{CMD} \cite{zellinger2017central} are typical DIRL methods for UDA. \textbf{DANN} leverages a domain discriminator to adversarially train the feature representation. \textbf{CMD} is a typical metric-based DIRL method which minimize a central moment discrepancy to measure the domain distribution shift. Both methods are implemented based on bert-base-cased model.

\textbf{EADA} \cite{zou-etal-2021-unsupervised} is a recently proposed DIRL method for cross-domain text classification task, which leverages an energy-based autoencoder to adversarially train the feature extractor.

\textbf{Multi-cell LSTM}\cite{jiazhang2020multi} A competitive cross-domain NER method that trains the source model in a multi-task manner with a language modeling task, an entity prediction task and an attention scoring task, which also requires lexicon-annotation. We implement this method using the same lexicon as ours.

\textbf{X-Pieces} The proposed method is implemented based on the bert-base-cased pre-trained model and default tokenizer. The lexicon we used is from \cite{liang2020bond}. For all experiments except that in Section {\ref{joint}}, we implemented based on optimizing the conditional distribution. We also combined our method with \textbf{DANN} for further improvement. We implement all models in MindSpore.

\subsection{Cross-domain Performance}
Table \ref{tab:ontonotes} and Table \ref{tab:conll} show the cross-domain results on setting (1) and (2), respectively. From the tables, we can observe that: 1) The proposed X-Piece method consistently exceeds BERT-tagger in most scenarios. In OntoNotes cross-domain setting, X-Piece achieves up to 4.58\% improvement over BERT-tagger when transferring from TC to MZ. As TC and MZ are two domains of the smallest training set (6.4k, 6.9k), we induce that X-Piece can bring more advantages on lower-resource cross-domain setting, where the data distribution shift causes more impact on the domain generalibility when trained with less data. 2) The feature-based methods (CMD, EADA) do not always benefit the performance when adapted to NER task, which verifies that transferring the high-dimensional features might not always be effective in word-level classification tasks like NER. 3) X-Piece can also benefit the performance when combined with the feature-based method (DANN), showing the effectiveness of the proposed method as a novel sub-word level domain adaptation tool.

\subsection{Correlation with KL-divergence}

\begin{figure}
    \centering
    \includegraphics[width=0.78\linewidth]{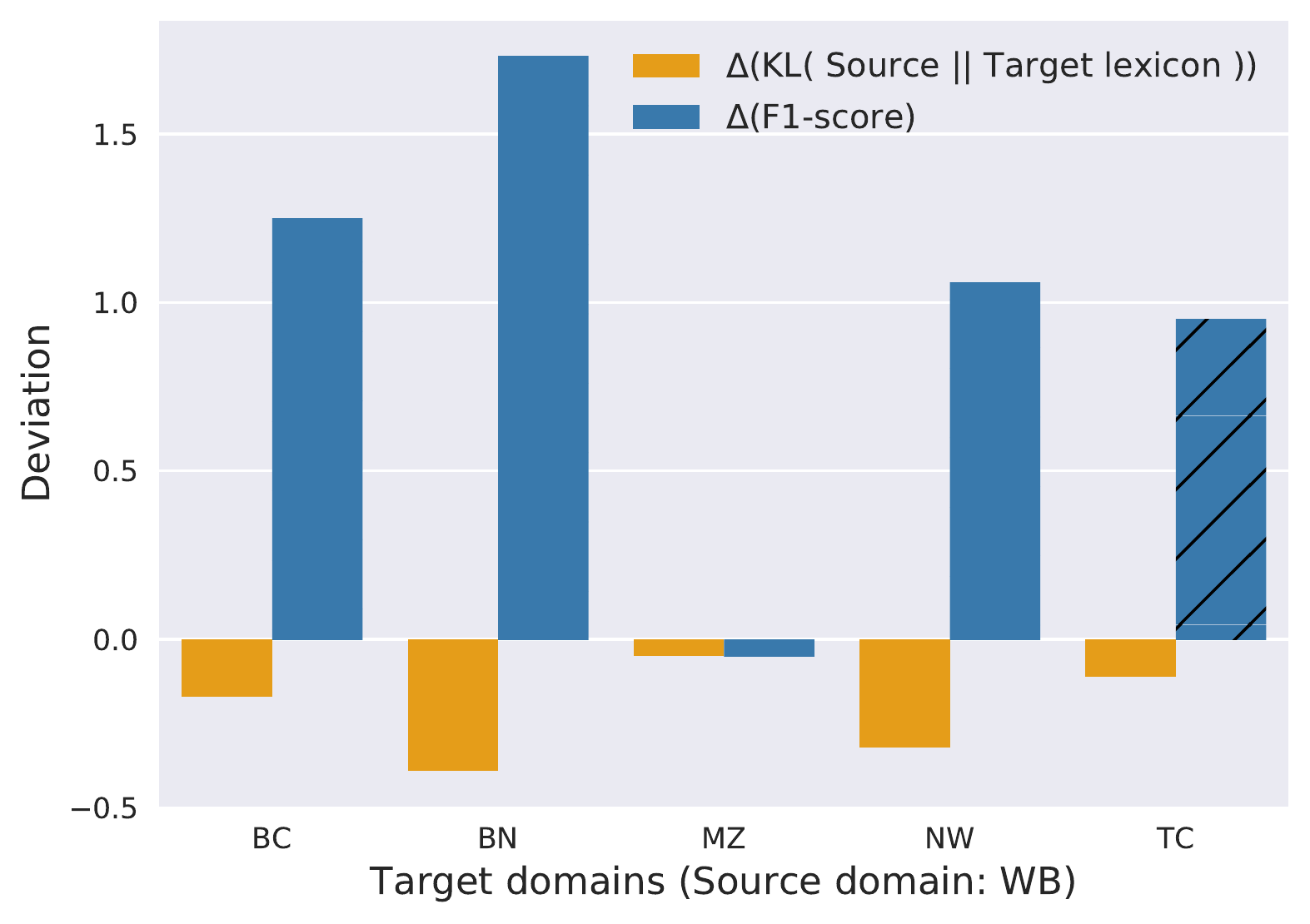}
    \caption{Deviations of the KL divergence score and performance gains before and after X-Piece. Higher KL-divergence degradation generally attributes to better performance.}
    \label{fig:kl2}
\end{figure}

As described in Section \ref{KLdivergence}, we use the KL-divergence score as a measure of the domain distribution shift, and optimize the subword distribution to minimize the KL score. How does the KL divergence vary after OT and how is the variation related to the performance? In Fig.\ref{fig:kl2}, we show the performance gain and the deviation of the KL-divergence before and after OT (WB to other OntoNotes sub-domains). Intuitively, we can observe that the KL-divergence between the source subword distribution and the target lexicon-approximated distribution consistently declines after optimized by OT. Also, the $|\Delta(KL)|$ shows a positive correlation with the performance gain, which verifies the rationality of using KL-divergence as the domain shift measure.

\subsection{Impact of Lexicon Annotation}

As mentioned in Section \ref{lexicon}, we use the lexicon annotation on the target domain unlabeled data for estimating the target distribution. In this section, we conduct experiments (WB to other OntoNotes sub-domains) to investigate the impact of lexicon noise, as shown in Tab.\ref{tab:lexicon_noise}. We train the BERT-tagger directly on the lexicon annotated data under two settings. The \textbf{further} setting first fine-tunes BERT-tagger on the source data and then further fine-tunes on the target lexicon annotated data; The \textbf{together} setting fine-tunes the model on the source data and target lexicon annotated data simultaneously. We can observe that the lexicon-annotated data can hardly benefits the model performance or even causes degradation when trained directly. While the proposed method, which only use the noisy data to estimate the target distribution, is relatively robust to the annotation noise. 

\begin{figure}
    \centering
    \includegraphics[width=0.78\linewidth]{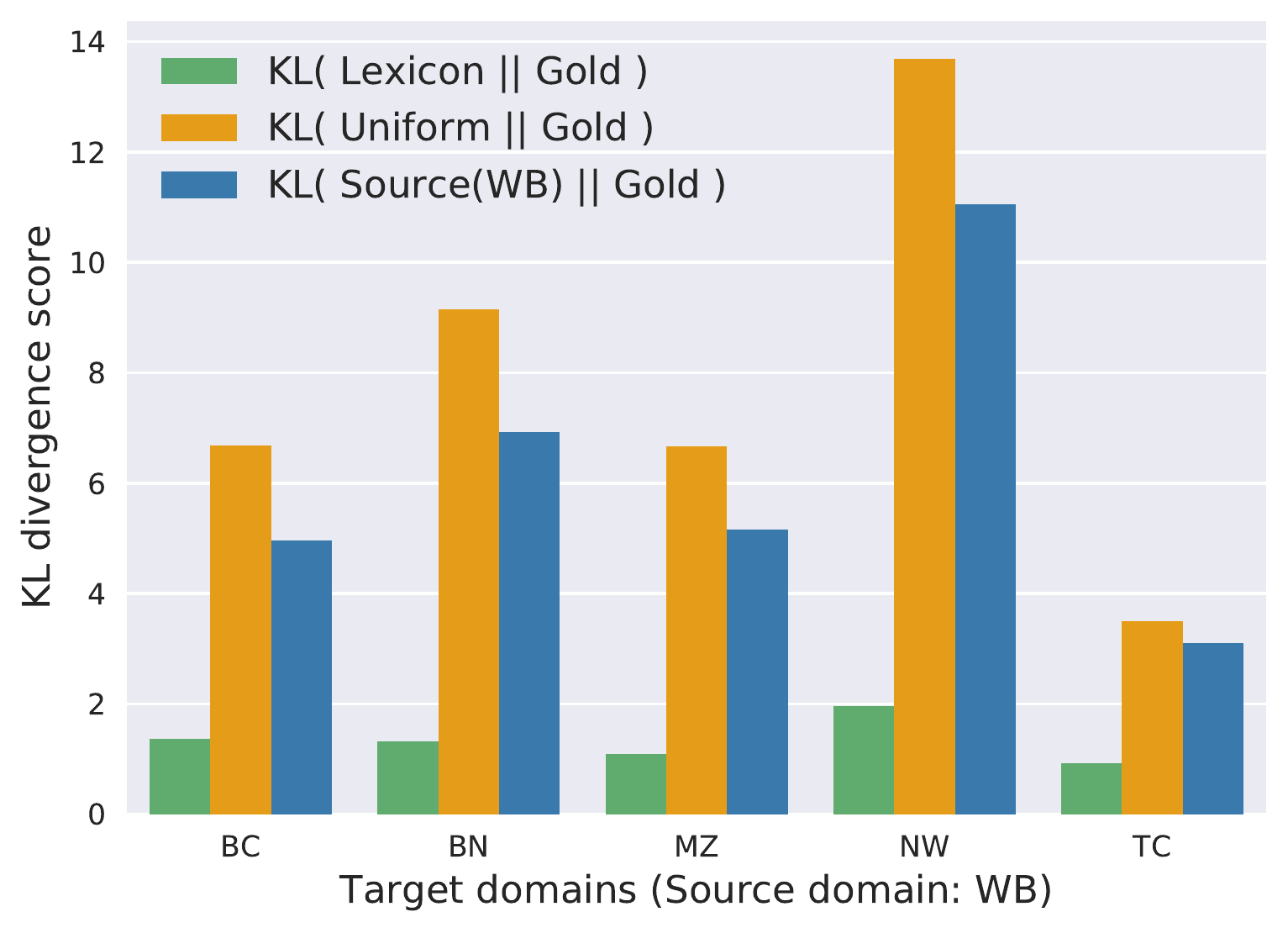}
    \caption{KL-divergence of the lexicon-approximated distribution against gold distribution (Lexicon || Gold), compared to other distributions. These results validate the usage of lexicon for estimating target distribution.}
    \label{fig:kl1}
    \vspace{-0.2cm}
\end{figure}

To further validate the proposed method, we also calculate the KL-divergence between the lexicon-approximated distribution and the gold-labeled distribution (Lexicon || Gold). As a comparison, we additionally show the KL-divergence between source distribution (Source || Gold), uniform distribution (Uniform || Gold) and gold-labeled distribution. The results are shown in Fig.\ref{fig:kl1}. We can observed that the KL-divergence between lexicon-approximated distribution and target distribution is lower than others, indicating that using the lexicon to estimate the target distribution is feasible.

\begin{table}[]
\centering
\small
\begin{tabular}{lccccc}
\toprule
\textbf{Methods} & BC & BN & MZ & NW & TC\\
\midrule
BERT-tagger & 69.60  & 77.06  & \textbf{78.51}  & 74.63 & 66.33  \\
 - further  & 65.55 & 76.42 & 77.54 & 74.79 & 66.02\\
 - together & 66.95 & 77.53 &76.95 & 74.90& {67.11}\\

X-Piece  &\textbf{70.85} & \textbf{78.79}  & 78.46 & \textbf{75.69} & \textbf{67.28}  \\

\bottomrule

\end{tabular}
\caption{The impact of the noisy lexicon-annotated data (Source domain: WB). While the noisy data is less useful or might cause performance degradation when directly training, our method is relatively robust to the data noise. }
\label{tab:lexicon_noise}
\vspace{-0.2cm}
\end{table}

\begin{figure}
\vspace{-0.3cm}
    \centering
    \includegraphics[width=0.7\linewidth]{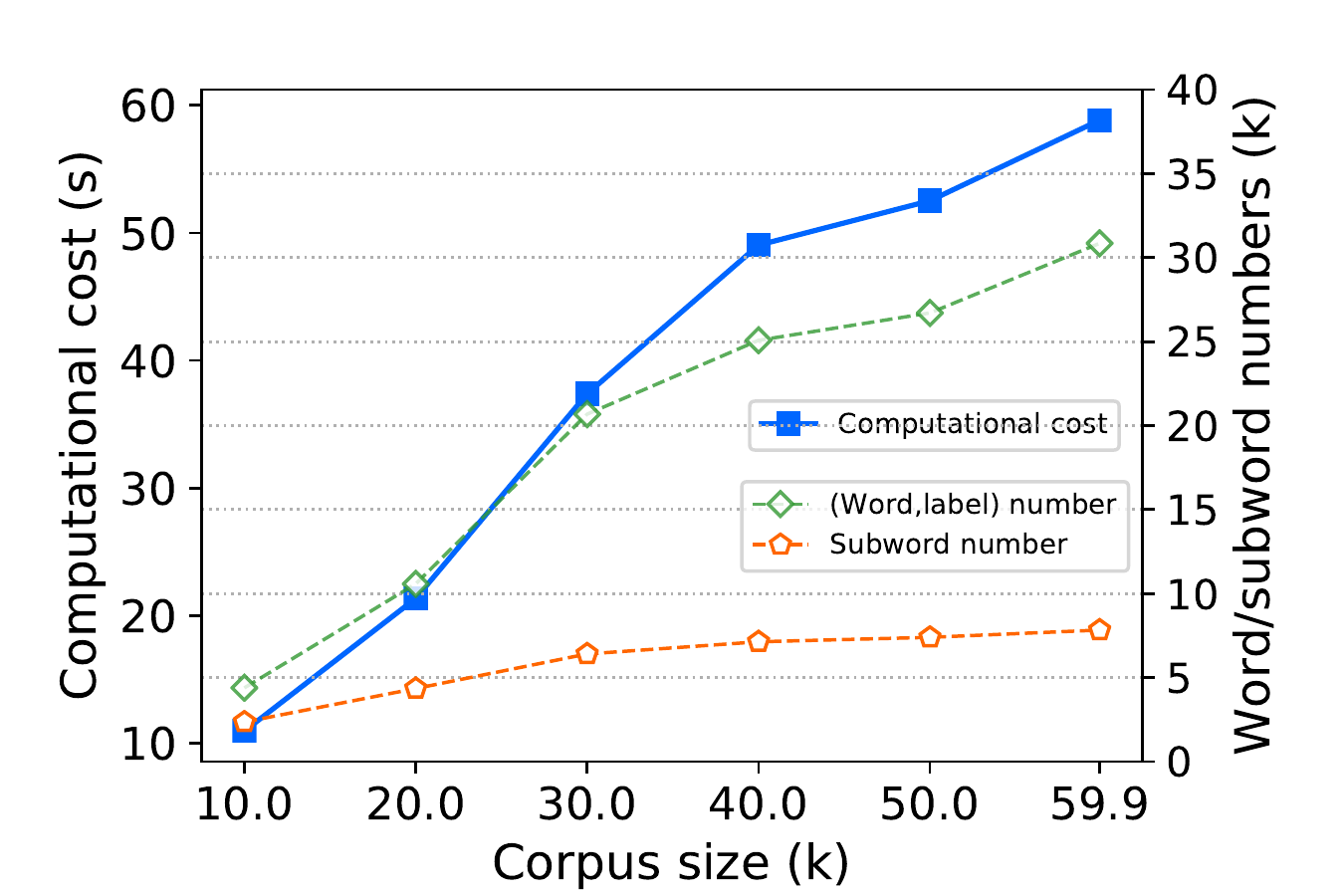}
    \caption{Computational cost of the pre-processing of X-Piece against corpus size. The computational cost is acceptable (up to $58.84s$) even with a distribution matrix size of $30.8k \times 7.8k$.}
    \label{fig:cost}
    \vspace{-0.2cm}
\end{figure}

\begin{table}[]
\centering
\small
\begin{tabular}{lccccc}
\toprule
\textbf{Methods} & BC & BN & MZ & NW & TC\\
\midrule
BERT-tagger & 69.60  & 77.06  & 78.51  & 74.63 & 66.33  \\
Ours (Joint) & 70.65 & 78.29 & \textbf{79.42} & 75.32 & \textbf{68.66}\\
Ours (Cond.)  & \textbf{70.85} & \textbf{78.79} & 78.46 & \textbf{75.69} & 67.28\\

\bottomrule

\end{tabular}
\caption{Comparison of different optimization objectives. "Ours (Joint)" denotes X-Piece using joint distribution as objective, "Ours (Cond.)" denotes X-Piece using conditional distribution as objective.}
\label{tab:joint}
 \vspace{-0.2cm}
\end{table}

\subsection{Computational Cost}
X-Piece contains a pre-processing process to calculate the distribution via OT, which can be done offline on a CPU machine. After one pass of pre-process, the distribution matrix can be stored for future tokenization. In this section, we conduct experiments to investigate the computational cost of the pre-processing part (OT) of X-Piece. Specifically, we test the computation time against different corpus size on Intel Xeon Platinum 8260, 2.40GHz. As shown in Fig.\ref{fig:cost}, the word number and the subword number increase as the corpus size rises, meaning that the distribution matrix to be solved is also expanding. As a result, the computation time of the proposed method also increases. However, the computational cost is acceptable (up to $58.84s$) even with a corpus size of 59.9k sentences and a distribution matrix size of $30.8k \times 7.8k$.

\subsection{Optimizing with Joint Distribution}\label{joint}
In Section \ref{formulate}, we formulate the optimal transport problem with two optimization objective, the conditional distribution ($P(\textbf{y}|\textbf{x})$) and the joint distribution ($P(\textbf{x},\textbf{y})$). The difference is that the conditional objective only cares about the relative distribution of labels given a certain subword, while optimizing the joint objective means additionally aligning the subword distributions. In Table \ref{tab:joint} we compare these two methods on WB to other OntoNotes sub-domains. 
Generally, the performance of the conditional objective is better than the joint objective. However, in some cases (WB to MZ, WB to TC) the joint objective show advantages. This might because the training set size of MZ and TC is relatively small, thus optimizing the joint distribution is feasible. In such cases, optimizing the joint distribution can better decrease the domain gap as it additionally solves the shift of $P(\mathbf{t})$.

\section{Conclusion}
In this work, we propose X-Piece, a subword-level approach for cross-domain NER, which alleviates the distribution shift between domains directly on the input data via subword-level distribution alignment. Specifically, we re-tokenize the input words in order to align the subword distribution with an lexicon-approximated target distribution. To find the optimal subword distribution, we formulate and solve it as an optimal transport problem. Experimental results verify the effectiveness of the proposed method on four NER datasets from different domains. Also, the proposed method is shown to benefit other cross-domain algorithms.

\section*{Acknowledgements}
The authors wish to thank the anonymous reviewers for their helpful comments. This work was partially funded by National Natural Science Foundation of China (No.  62076069,61976056), Shanghai Municipal Science and Technology Major Project (No.2021SHZDZX0103). This research was supported by Meituan and CAAI-Huawei MindSpore Open Fund.

\bibliographystyle{named}
\bibliography{custom}

\appendix

\section{Sinkhorn Algorithm}
In our experiments, we solve the optimal source domain distribution by solving the formulated OT problem through the Sinkhorn algorithm. The details of our implementation is shown in Algorithm \ref{alg:ot}.

\section{Derivation of formulating the Optimal Transport (OT) Problem}
In this section, we detail the derivation of formulating the OT problem in Section 3.2.
We first derive the formulation based on optimizing the conditional distribution $P(\textbf{y}|\textbf{t})$. The KL-divergence between the source and target conditional distribution is given by:

\begin{equation}\label{KL}
\setlength\abovedisplayskip{0pt}
\setlength\belowdisplayskip{2pt}
\begin{aligned}
D_{KL}&(P_S(\textbf{y}|\textbf{t})\|P_T(\textbf{y}|\textbf{t})) \\
&=\underbrace{\sum_{t\in\mathbb{T}} \sum_{y\in\mathbb{Y}} P_S(y|t)\log(P_S(y|t))}_{\mathcal{L}_1} \\
&  \underbrace{-\sum_{t\in\mathbb{T}} \sum_{y\in\mathbb{Y}}P_S(y|t)\log(P_T(y|t)) }_{\mathcal{L}_2}\\
\end{aligned}
\end{equation}
where $\mathcal{L}_1$ is the negative entropy $-\mathcal{H}(P_S(\textbf{y}|\textbf{t}))$ of the source domain conditional distribution. 

Notice that the above formulation haven't considered the word distribution $P_S(\textbf{w})$. However, we cannot ignore the word distribution since the input word distribution is fixed and the optimization of subword distribution is limited by the given words. Therefore, we re-formulate the term $P_S(\textbf{y}|\textbf{t})$ into:
\begin{equation}
\setlength\abovedisplayskip{3pt}
\setlength\belowdisplayskip{0pt}
\begin{aligned}
P_S(\textbf{y}|\textbf{t}) &= \sum_{w\in\mathbb{V}}P_S(\textbf{y},w|\textbf{t}) 
&= \sum_{w\in\mathbb{V}} \frac{P_S(\textbf{t},w,\textbf{y})}{P_S(\textbf{t})}
\end{aligned}
\end{equation}

where $\mathbb{V}$ is the word vocabulary. Then, we can reformulate the $\mathcal{L}_2$ term in Eq.\ref{KL} as:
\begin{equation}
\setlength\abovedisplayskip{4pt}
\setlength\belowdisplayskip{0pt}
\begin{aligned}
\mathcal{L}_2 = -&\sum_{t\in\mathbb{T}} \sum_{y\in\mathbb{Y}} \sum_{w\in\mathbb{V}} \frac{P_S(t,w,y)}{P_S(t)}log(\frac{P_T(t,y)}{P_T(t)}) \\
=&\sum_{t\in\mathbb{T}} \sum_{(w,y)\in\mathbb{V}\times\mathbb{Y}} 
[P_S(t,w,y)\\
&\cdot-\frac{1}{P_S(t)}log(\frac{P_T(t,y)}{P_T(t)})]
\end{aligned}
\end{equation}
where $(w,y)$ is the joint term of a certain word $w$ and a certain label $y$.

Here, the joint distribution $P_S(t,w,y)$ is the objective to solve via the optimization, which corresponds to the joint distribution matrix $\mathbf{P}$ in the optimal transport. The term $-\frac{1}{P_S(t)}log(\frac{P_T(t,y)}{P_T(t)})$ can be regarded as the transport cost of transporting the subword $t$ into the word-label term $(w,y)$, corresponding to the $(|\mathbb{V}|\times|\mathbb{Y}|)\times|\mathbb{T}|$ cost matrix $\mathbf{D}$ in OT. To calculate this cost term, the $P_T(t,y)$ and $P_T(t)$ terms can be collected from the target domain, which is detailed in Sec.\ref{lexicon}. However, the ${P_S(t)}$ is the unknown distribution that we aim to solve, which cannot be obtained beforehand. Therefore, we turn to minimize the upper bound of $\mathcal{L}_2$ based on $\frac{1}{P_S(t)}\geq 1$:
\begin{equation}
\setlength\abovedisplayskip{3pt}
\setlength\belowdisplayskip{0pt}
\begin{aligned}
\mathcal{L}_2
\leq \overline{\mathcal{L}_2} = &\sum_{t\in\mathbb{T}} \sum_{(w,y)\in\mathbb{V}\times\mathbb{Y}} 
[P_S(t,w,y)\\
&\cdot-log(\frac{P_T(t,y)}{P_T(t)})]
\end{aligned}
\end{equation}

As a result, we have gotten a similar form with the optimal transport problem:
\begin{equation}
\setlength\abovedisplayskip{4pt}
\setlength\belowdisplayskip{1pt}
\begin{aligned}
\mathcal{L}_2&=<\mathbf{P},\mathbf{D}> \\
&=\sum_{t\in\mathbb{T}} \sum_{(w,y)\in\mathbb{V}\times\mathbb{Y}} \mathbf{P}(t,w,y)\mathbf{D}(t,w,y)
\end{aligned}
\end{equation}

In this way, Eq.\ref{KL} can be reformulated as the following objective function which has the same form as the objective function in optimal transport:
\begin{equation}\label{OT_final}
\setlength\abovedisplayskip{0pt}
\setlength\belowdisplayskip{0pt}
\begin{aligned}
\min_{\mathbf{P}\in \mathcal{R}^{(w,y)\times t}}<\mathbf{P},\mathbf{D}>-\gamma \mathcal{H}(\mathbf{P})
\end{aligned}
\end{equation}

Additionally, we also can formulate the OT problem based on optimizing the joint distribution $P(\textbf{t},\textbf{y})$, which has a similar form with Eq.\ref{OT_final} despite that each term of the cost matrix \textbf{D} is $-log(P_T(t,y))$.

\begin{algorithm}[t]
\footnotesize
\caption{Sinkhorn Algorithm}
\textbf{Input:} A gold-labeled source-domain corpus $D_{S}$, a target-domain corpus $D_{T}$ with distant annotations labeled by entity lexicon, an identical subword space $\mathbb{T}$, an identical entity category space $\mathbb{Y}$, the word vocabulary $\mathbb{W}$   \\
\textbf{Parameters:} $\mathbf{u}$ $\in \mathbb{R}_+^{|\mathbb{C}|}$, $\mathbf{v}$ $\in \mathbb{R}_+^{|\mathbb{T}|}$   \\
\tcp{Begin of Sinkhorn algorithm }
Initialize $\mathbf{u}$ = ones() and $\mathbf{v}$ = ones() \\

 Calculate the joint distrubution $P_S(\mathbb{W},\mathbb{Y})$ based on $D_{S}$ \\
Calculate the subword distribution $P_S(\mathbb{T})$ based on $D_{S}$ \\
Calculate the cost matrix $\mathbf{D}$ based on $D_{T}$ \\
\While{not converge}{
\textbf{u} = $ P_S(\mathbb{W},\mathbb{Y}) / \mathbf{D}\textrm{v} $\\
\textbf{v} = $P_S(\mathbb{T}) / \mathbf{D}^T \textrm{u} $
}
optimal\_matrix $\mathbf{P}$ = $\mathbf{u}$.reshape(-1, 1) * $\mathbf{D}$ * 
$\mathbf{v}$.reshape(1, -1) \\
\tcp{End of Sinkhorn algorithm } 
Output optimal\_matrix $\mathbf{P}$

\label{alg:ot}
\end{algorithm}

\end{document}